\documentclass{article}

\usepackage{arxiv}
\usepackage[utf8]{inputenc} 
\usepackage[T1]{fontenc}    
\usepackage[hidelinks]{hyperref}       
\usepackage{url}            
\usepackage{booktabs}       
\usepackage{amsfonts}       
\usepackage{nicefrac}       
\usepackage{microtype}      
\usepackage{lipsum}		
\usepackage{graphicx}
\usepackage[authoryear,round]{natbib}
\usepackage[english]{babel}
\usepackage[hang,flushmargin]{footmisc}

\newcommand{\ncite}[1]{%
  \citeauthor{#1},~\citeyear{#1}%
}

\title{Visual Image Reconstruction from Brain Activity via Latent Representation}

\author{ 
        {
        Yukiyasu Kamitani}
        \\
	Kyoto University and \\ 
        ATR Computational Neuroscience Laboratories \\
        Kyoto, Japan \\
	\texttt{kamitani@i.kyoto-u.ac.jp} \\
	\And
        {
        Misato Tanaka} \\
	Kyoto University and \\ 
        ATR Computational Neuroscience Laboratories \\
        Kyoto, Japan \\
	\texttt{mtanaka@i.kyoto-u.ac.jp} \\
	\And
        {
        Ken Shirakawa} \\
	ATR Computational Neuroscience Laboratories\\
        Kyoto, Japan \\
	\texttt{shirakawaken0118@gmail.com} \\
}

\renewcommand{\footnotesize}{\small} 

\renewcommand{\shorttitle}{Visual Image Reconstruction from Brain Activity via Latent Representation}

\hypersetup{
pdftitle={Visual Image Reconstruction from Brain Activity via Latent Representation},
pdfsubject={q-bio.NC, cs.CV, cs.LG}, 
pdfauthor={Yukiyasu Kamitani, Misato Tanaka, Ken Shirakawa},
pdfkeywords={visual image reconstruction, brain decoding, deep neural network (DNN), latent representation, zero-shot prediction, NeuroAI},
}

\begin{document}
\maketitle

\begin{abstract}
Visual image reconstruction, the decoding of perceptual content from brain activity into images, has advanced significantly with the integration of deep neural networks (DNNs) and generative models. This review traces the field’s evolution from early classification approaches to sophisticated reconstructions that capture detailed, subjective visual experiences, emphasizing the roles of hierarchical latent representations, compositional strategies, and modular architectures. Despite notable progress, challenges remain, such as achieving true zero-shot generalization for unseen images and accurately modeling the complex, subjective aspects of perception. We discuss the need for diverse datasets, refined evaluation metrics aligned with human perceptual judgments, and compositional representations that strengthen model robustness and generalizability. Ethical issues, including privacy, consent, and potential misuse, are underscored as critical considerations for responsible development. Visual image reconstruction offers promising insights into neural coding and enables new psychological measurements of visual experiences, with applications spanning clinical diagnostics and brain–machine interfaces.
\end{abstract}

\keywords{visual image reconstruction \and brain decoding \and deep neural network (DNN) \and latent representation \and zero-shot prediction \and NeuroAI}

\section{Introduction}
\begin{figure}[t]
\centering
\includegraphics[scale=0.64]{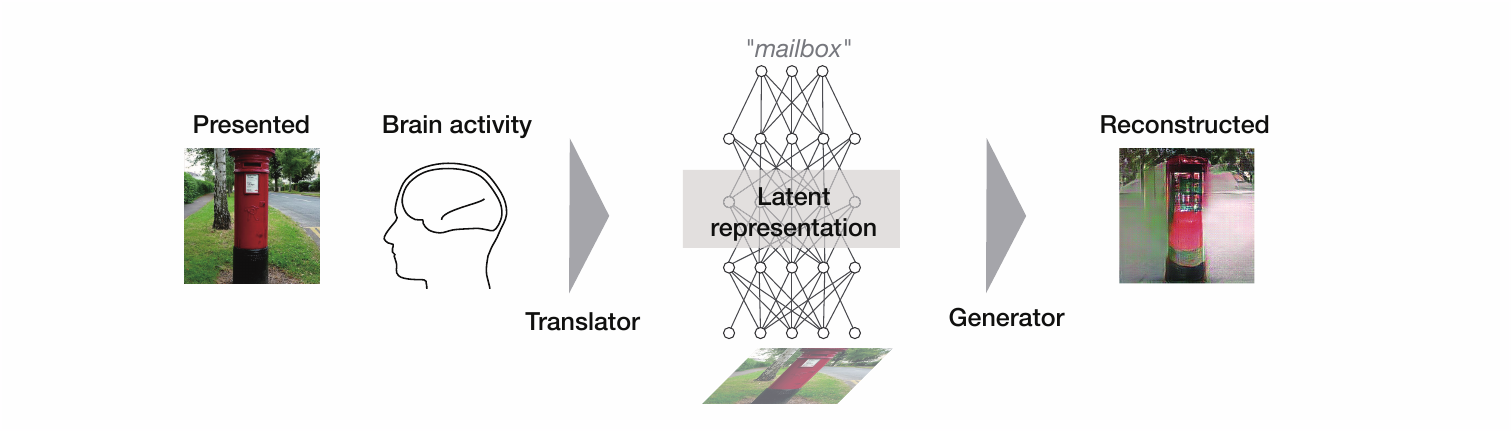}
\caption{General framework for visual image reconstruction. The process begins by translating brain activity patterns, recorded as a person views an image, into a latent representation within a machine vision model, typically a deep neural network (DNN). This step, termed {\it translation} here, treats brain activity as a latent representation of the observed image, emphasizing the translation between neural and machine representations. In the second step, a generator module uses these decoded latent features to reconstruct a visual image corresponding to the neural representation, often employing generative AI methods.}
\label{fig:image_reconstruction_pipeline}
\end{figure}

Visual image reconstruction\footnote{\textbf{Visual image reconstruction:} The process of decoding perceptual content from brain activity as images that approximate visual experiences.}, a rapidly advancing technique for decoding perceptual content from brain activity, has undergone significant improvements in recent years. Early efforts leveraged linear filtering properties and the retinotopic organization in the early visual pathway (\ncite{stanley_reconstruction_1999}; \ncite{thirion_inverse_2006}), leading to the reconstruction of perceptible images from human functional magnetic resonance imaging (fMRI) in V1 using a combination of low-level image bases (\ncite{miyawaki_visual_2008}). However, the advent of deep learning methods has revolutionized this field, enabling the generation of higher-quality images that more accurately represent the detailed content of perception (\ncite{shen_deep_2019}; \ncite{seeliger_generative_2018}). Harnessing hierarchical latent representations of visual images derived from deep neural networks\footnote{\textbf{Deep neural network (DNN):} Artificial intelligence models that process data through multiple interconnected layers to learn hierarchical representations of complex features.} (DNNs; \ncite{horikawa_generic_2017}), these approaches now offer a pathway to reconstruct more complex and subjective visual experiences (\ncite{shen_deep_2019}; \ncite{shen_end--end_2019}; \ncite{horikawa_attention_2022}; \ncite{cheng_reconstructing_2023}; Figure \ref{fig:image_reconstruction_pipeline}). 

Visual image reconstruction aims to generate images that accurately replicate the subjective perceptual experiences encoded in the brain beyond simply mirroring external stimuli. This method has profound implications across neuroscience and applied technology. First, it provides a novel method for understanding neural coding by assessing the information extractable from brain activity, thereby serving as a robust test of neural encoding. Unlike the decoding of isolated features, whole-image reconstruction is critical for understanding how neural encoding impacts holistic visual experiences (\ncite{cheng_reconstructing_2023}).

Second, visual reconstruction introduces novel methods for psychological measurement\footnote{\textbf{Psychological measurement:} The systematic quantification of mental processes through measurable variables.}, which systematically quantify mental processes by assigning measurable variables. Traditional psychophysics has developed rigorous methodologies to quantify perceptual content, but it often focuses on measuring specific variables using simple behavioral responses such as button presses. Similarly, while neuroimaging has identified {\it biomarkers} for psychological processes, it has generally only captured coarse representations of mental states. In contrast, image reconstruction offers a more direct approach by externalizing internal mental states as detailed images composed of pixel values. This capability transforms mental phenomena into tangible visual outputs, allowing for a more precise and comprehensive analysis of psychological states and processes.

Finally, visual image reconstruction could hold significant promise for applications in clinical diagnostics and brain-machine/computer interfaces (BMIs/BCIs). In clinical diagnostics, visual reconstruction could assist in assessing hallucinatory experiences associated with neurological and psychiatric disorders. By reconstructing visual content based on neural activity, clinicians may gain insights into the nature and origins of hallucinations, providing a non-invasive tool for diagnosis, monitoring, and potentially guiding therapeutic interventions.

Traditionally, BMIs have focused on motor control for tasks such as operating prosthetics or robotic limbs (\ncite{hochberg_reach_2012}). Integrating visual reconstruction, however, could expand BMIs to enable the communication of visual content, creating new non-verbal interaction pathways for individuals with severe physical impairments. Nonetheless, these advancements raise important ethical considerations, including issues of privacy, informed consent, and the implications of accessing and interpreting subjective mental imagery.

This article provides an overview of visual image reconstruction, tracing its historical development, evaluating its current capabilities, and exploring its future potential. By examining key technical aspects and theoretical implications, we aim to show how brain decoding\footnote{\textbf{Brain decoding:} The prediction of mental content from brain activity using machine learning, often applied to non-invasive neuroimaging data.} combined with DNN-based latent representations\footnote{\textbf{Latent representation:} A set of underlying features that encapsulate information about observed content, and serve as the basis for generation.} can deepen our understanding of visual perception and guide future brain-based technologies. Recent reviews provide complementary perspectives on the promises and pitfalls of brain decoding and DNNs for modeling human vision. \cite{robinson_visual_2023} illustrate how decoding techniques can uncover internal visual representations, while \cite{wichmann_are_2023} question whether current deep architectures genuinely capture the richness and variability of human vision. These discussions align with our analysis, emphasizing the importance of carefully selecting and critically evaluating AI models. Although we focus primarily on fMRI-based visual image reconstruction, alternative decoding methods using EEG/MEG further highlight the flexibility and richness of neural signals, offering a broader methodological perspective on how to map perceptual features from brain activity.

\section{Historical context: Decoding and reconstruction}
The foundations of visual image reconstruction are rooted in early efforts to decode sensory information from neuronal signals, building on the principles of neural coding. Bialek and colleagues were pioneers in this field, defining neural signals as those that could be most accurately reconstructed by observing spike trains of specific neurons (\ncite{bialek_reading_1991}). Their work provided a framework for estimating time-varying stimuli from neural activity, demonstrating that stimulus waveforms could be reconstructed with high accuracy using linear filters. Their methodology has been applied across sensory systems, demonstrating the feasibility of stimulus reconstruction in different modalities (\ncite{rieke_spikes:_1997}).

Another critical development in neural decoding was the introduction of population vectors, initially developed to predict movement direction from motor cortex activity (\ncite{georgopoulos_neuronal_1986}; \ncite{salinas_vector_1994}). This method proposed that the collective activity of a neural population could be linearly combined to generate a vector pointing to the intended movement direction. It forms the foundation for brain–machine interfaces (BMIs), enabling robotic arm control through linear readout\footnote{\textbf{Linear readout:} An operation that linearly maps neural activation patterns to outputs, determining if information is explicitly encoded within the pattern.} techniques (\ncite{schwartz_extraction_2001}). While primarily focused on motor control, these principles were later extended to sensory decoding, such as the decoding of visual motion directions.

Applying these concepts to visual information presented new challenges, as visual signals are inherently more complex than single-dimensional signals or movement directions. Early work by \cite{stanley_reconstruction_1999} attempted to decode visual information from the lateral geniculate nucleus (LGN) of cats. By training linear filters to translate neuronal population activity into pixel values, inverse to the receptive field modeling, this work demonstrated the potential to reconstruct sensory information from early visual processing centers that encode pixel values linearly.

An integrative approach to neural signal decoding frames stimulus estimation as a statistical problem, with neural responses acting as observations and unknown stimulus parameters inferred through maximum likelihood or Bayesian methods (\ncite{paradiso_theory_1988}; \ncite{seung_simple_1993}; \ncite{dayan_theoretical_2001}). In this framework, neural responses are typically modeled by Gaussian or cosine-like tuning functions with added noise. This formulation enables maximum likelihood estimates of stimulus parameters to be expressed as a weighted sum of each neuron's preferred stimulus parameters, where the weights correspond to neural activity levels, consistent with linear readout principles. Such a framework captures the probabilistic nature of neural responses and underscores the efficiency of linear decoding strategies for extracting sensory information.

The Bayesian perspective has led to the development of an approach in which sensory encoding models are trained from data and then inverted for decoding (\ncite{naselaris_encoding_2011}). An early attempt at visual reconstruction involved inverting retinotopy in the early visual cortex. Retinotopy refers to the spatial organization of neural responses that reflect the layout of the retina, making it a natural strategy for visual image reconstruction. \cite{thirion_inverse_2006} employed fMRI to invert retinotopic maps, revealing some visual stimulus information, though the resulting images were not readily perceptible. Later studies (\ncite{naselaris_bayesian_2009}; \ncite{schoenmakers_linear_2013}) utilized more sophisticated methods to design and invert encoding models, which improved reconstruction performance, although substantial limitations remained.

Reconstruction typically implies the capacity to predict continuous, arbitrary values within the stimulus parameter space, rather than simply classifying inputs into predefined categories. Consequently, for large stimulus spaces that cannot be fully sampled during training, an essential requirement for effective reconstruction is the model's ability to generalize to novel instances beyond those seen in the training set. However, some early studies do not appear to test this aspect critically, leaving questions about the models’ generalization capabilities. With the rise of modern machine learning, awareness of this issue has grown, emphasizing the need for models that can reliably generalize to unseen content.

\section{Brain decoding: From classification to reconstruction}
While neural encoding/decoding research in neurophysiology has primarily relied on spike data, another origin of visual image reconstruction lies in neuroimaging studies (\ncite{kamitani_decoding_2012}). In fMRI research, the major focus has been on functional brain mapping, aiming to localize activity related to some coarse cognitive functions, such as memory and language, to specific brain regions. This localization-based approach typically analyzed average activity within regions of interest, often missing finer patterns of neural representation. A pivotal shift occurred with the study by \cite{haxby_distributed_2001}, which emphasized distributed patterns of activation in the ventral visual cortex to discriminate different object categories, moving beyond simple localization. \cite{cox_functional_2003} applied machine learning classifiers to the same data to achieve reliable predictions.

Despite these advances, studying basic visual features such as orientation and color using fMRI posed significant challenges due to the fine-scale neural representations identified in animal studies and the limited spatial resolution of fMRI. For example, visual orientation is believed to be represented through columnar organization (\ncite{hubel_receptive_1968}), which lies below the resolution of standard fMRI voxels. \cite{kamitani_decoding_2005} introduced a novel strategy to leverage the weak biases of individual voxels toward specific orientations. They hypothesized that random variations in neural distribution across voxels could be harnessed by pooling responses from multiple voxels, amplifying these biases and thereby enhancing effective orientation selectivity.

\cite{kamitani_decoding_2005} built a machine learning-based {\it orientation decoder} capable of accurately predicting the orientation of grating stimuli based on ensemble fMRI activity in early visual areas such as V1 and V2. While individual voxels exhibited low selectivity, the combined responses of multiple voxels generated sharp orientation tuning, allowing for robust and stable decoding across sessions. This series of work marked the emergence of brain decoding, a novel approach that leverages machine learning to analyze various brain signals, including non-invasive neuroimaging, to infer mental content from subtle, imperceptible patterns—often lacking apparent tuning seen in neurophysiological data (\ncite{kamitani_decoding_2006}; \ncite{kamitani_decoding_2012}).

Despite successful replications of orientation prediction, the exact source of this information remains a topic of debate (\ncite{op_de_beeck_against_2010}; \ncite{kamitani_spatial_2010}; \ncite{kriegeskorte_how_2010}; \ncite{swisher_multiscale_2010}; \ncite{roth_natural_2022}). A key insight from this research is that focusing solely on human-interpretable neural patterns risks overlooking valuable signals. The brain’s organization is complex and not necessarily structured for straightforward human interpretation. In studying complex systems, predictive accuracy often takes precedence over explanatory clarity (\ncite{yarkoni_choosing_2017}; \ncite{nastase_keep_2020}; \ncite{hasson_direct_2020}). As we will discuss, hierarchical representations in the brain and in DNNs frequently lack intuitive, interpretable labels, making them {\it nameless} and {\it faceless}. However, predictive modeling approaches have demonstrated success even in the absence of directly interpretable signals.

Another key aspect of \cite{kamitani_decoding_2005} is the concept of {\it neural mind-reading}, where a decoder trained on brain activity evoked by stimuli (e.g., single gratings) can be used to predict the subjective content (e.g., the attended orientation on overlapping gratings). This approach assumes a shared representation of the same content between perception and subjective states, such as attention, mental imagery, and dreaming (\ncite{stokes_top-down_2009}; \ncite{harrison_decoding_2009}; \ncite{horikawa_neural_2013}). The generalization of decoding models, or {\it cross-decoding}\footnote{\textbf{Cross-decoding:} Applying a model trained on one condition to decode another, testing shared neural representations across tasks or conditions.}, has become a fundamental technique for studying shared neural representations. It also plays a critical role in the visual image reconstruction of subjective experiences.

\begin{figure}[t]
\centering
\includegraphics[scale=0.64]{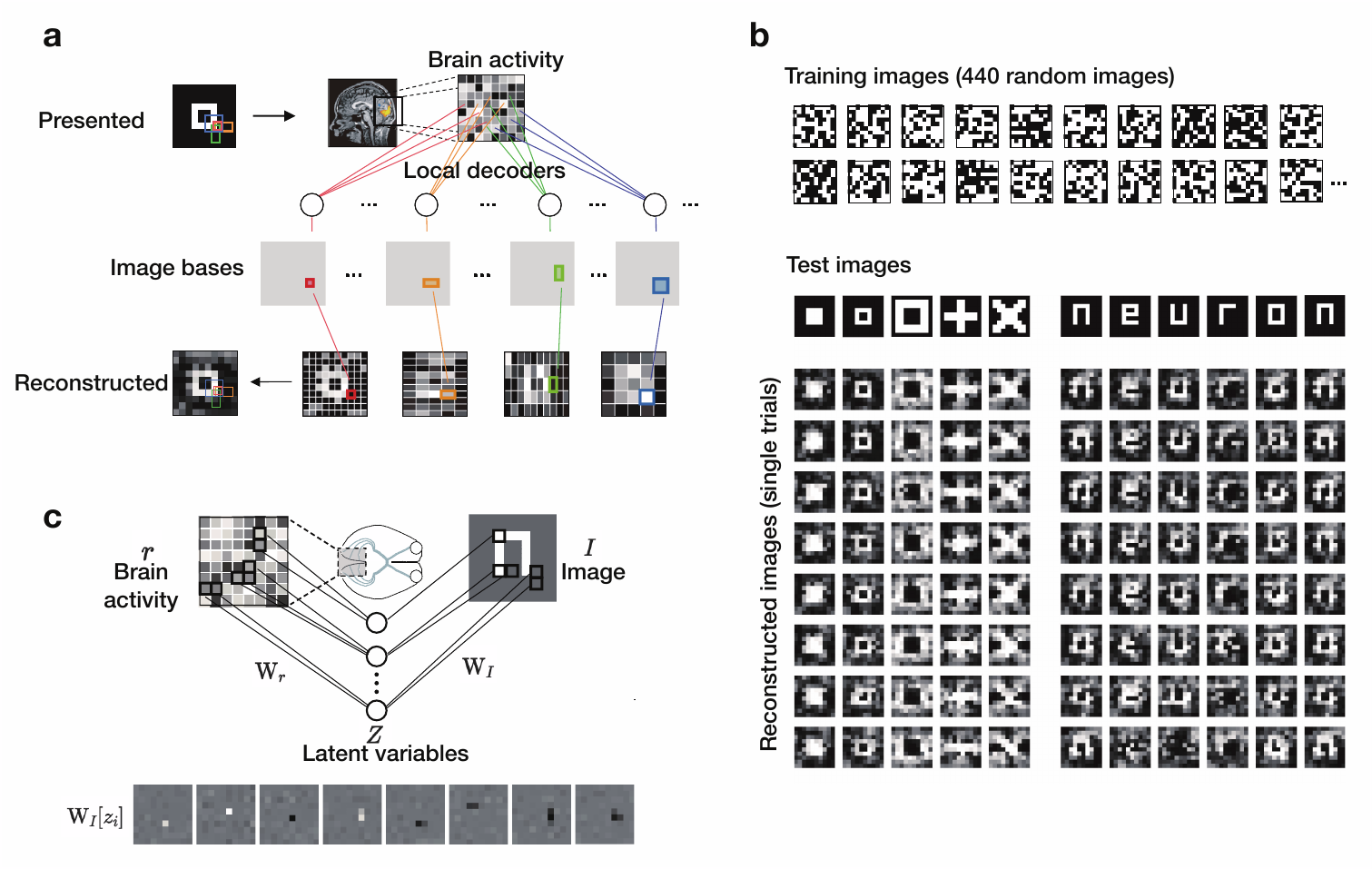}
\caption{Visual image reconstruction through a linear combination of image bases. (a) Reconstruction method by \cite{miyawaki_visual_2008}. fMRI activity is recorded as subjects view a 10 x 10 binary contrast image. {\it Local decoders} use linearly weighted multivoxel fMRI signals to predict the mean contrasts of local image bases at multiple scales, capturing a localized representation in the early visual cortex. Predicted contrasts are multiplied by the corresponding image bases and linearly combined to reconstruct the image. V1 voxels serve as input for each local basis, with relevant voxels selected through the method of \cite{yamashita_sparse_2008}, establishing a modular decoding approach. (b) Training and testing. Local decoders are trained on 440 random images, with test images including artificial shapes and patterns outside the training set. Single-trial reconstructions of shapes and letters demonstrate the model’s accuracy. (c) Bayesian canonical correlation analysis (BCCA) method by \cite{fujiwara_modular_2013}. This approach estimates local image bases by linking image pixels with fMRI activity patterns through latent variables with sparse weights. The network is trained using Bayesian estimation with sparseness priors, and the resulting image bases are shown below. Adapted from \cite{miyawaki_visual_2008} and \cite{fujiwara_modular_2013}.}
\label{fig:image_reconstruction_imagebase}
\end{figure}

\cite{miyawaki_visual_2008} advanced brain decoding research by developing a method for visual image reconstruction. Single-feature prediction approaches fall short of capturing the full complexity of perceptual experience, as perception encompasses an enormous range of potential states. For example, even a 10 x 10 binary image can yield up to 2100 possible configurations, making it impractical to measure brain activity for all such states. To overcome this limitation, \cite{miyawaki_visual_2008} proposed a modular decoding approach that reconstructs visual images as combinations of multiscale local image bases (Figure \ref{fig:image_reconstruction_imagebase}a), following the principles outlined by \cite{olshausen_emergence_1996}.

In this model, a decoder predicts the stimulus state for each local image element based on multivoxel fMRI patterns. The outputs of these local decoders are then combined in a statistically optimal way to reconstruct the overall image. Because each local element has fewer possible states than the entire image, this approach requires fewer training samples for effective decoding. The model draws on all available voxels from early visual areas, using a sparse estimation algorithm to automatically prune irrelevant voxels (\ncite{yamashita_sparse_2008}). Each of the resulting local decoders functions {\it as a module} that represents a fundamental image component with a small number of voxels, and the combination of these modules enables the representation of a wide range of complex images.

\cite{miyawaki_visual_2008} tested this approach by reconstructing images composed of 10 x 10 binary contrast-defined patches (pixels) (Figure \ref{fig:image_reconstruction_imagebase}b). After training the model with 440 random images, it reconstructed a variety of images, including geometric shapes and alphabetic characters on a single trial (6-s/12-s block) or single volume (2-s) basis without prior knowledge of the images. 

While \cite{miyawaki_visual_2008} used fixed image bases, \cite{fujiwara_modular_2013} introduced a method to automatically learn image bases from data (Figure \ref{fig:image_reconstruction_imagebase}c). Their approach used latent variables to establish relationships between image pixels and fMRI voxels, allowing for predictions in both encoding and decoding directions. This method employed canonical correlation analysis (CCA), which identifies correspondences, via latent variables, between weighted sums of pixels and voxels. To improve the mapping between small sets of pixels and voxels, Fujiwara et al. extended CCA to Bayesian CCA, incorporating sparseness priors for both pixel and voxel weights. In this framework, pixel weights for each latent variable define image bases, which are learned directly from the data. Applied to the dataset from \cite{miyawaki_visual_2008}, this model automatically identified spatially localized image bases, many resembling those used in the earlier model.

While these methods resemble the linear filtering approach used to predict individual pixel values (\ncite{stanley_reconstruction_1999}), incorporating image bases and latent representations laid the groundwork for using hierarchical latent features in visual image reconstruction. An important aspect of this early work was its modular architecture, which allowed for {\it zero-shot prediction}\footnote{\textbf{Zero-shot prediction:} The ability of a model to accurately reconstruct or identify novel images or information it has not encountered during training, demonstrating generalization beyond its training set.} of novel content through {\it compositional representations}\footnote{\textbf{Compositional representation:} A representation that treats complex content, such as images, as combinations of simpler elements, facilitating flexible and diverse reconstruction.} of stimuli, even with limited training samples.

In machine learning, zero-shot prediction refers to a model’s ability to accurately handle novel content not encountered during training. This concept aligns with brain decoding techniques that interpret neural patterns associated with previously unseen stimuli, allowing models to generalize from limited training data. To address the challenge of predicting novel stimuli from neural signals, researchers have developed various decoding methods. For example, \cite{kay_identifying_2008} proposed a statistical encoding model that predicts fMRI voxel responses from image features; by synthesizing brain activity corresponding to different feature combinations, they successfully identified novel test images from a set of candidates.  Similarly, \cite{mitchell_predicting_2008} demonstrated zero-shot identification of unseen nouns from brain activity by using verb co-occurrence rates as features in an encoding model. These approaches also rely on compositional representations of stimuli, allowing models to leverage combinations of previously learned features, thus enabling novel interpretations without prior exposure to specific examples.

The compositional capability in modular decoding models, essential for zero-shot prediction, is often overlooked in recent generative AI-based reconstruction methods, which tend to produce spurious outputs as discussed later (\ncite{shirakawa_spurious_2024}). By learning the underlying structure and relationships among elements, models employing modular mappings between input and output spaces can generalize to new instances, offering a more robust and interpretable framework for image reconstruction.

\section{NeuroAI}
DNNs have revolutionized artificial intelligence and neuroscience, offering powerful tools for modeling brain function (\ncite{kriegeskorte_deep_2015}).  The concept of hierarchical organization in DNNs is inspired by Hubel and Wiesel's pioneering work on the visual system, which identified {\it simple cells} and {\it complex cells} that form the basis for processing visual information at different levels of abstraction (\ncite{hubel_receptive_1968}). Building on these ideas, \cite{fukushima_neocognitron:_1980} developed the Neocognitron, which used alternating simple and complex layers corresponding to convolution and pooling layers in modern convolutional neural networks (CNNs).

This hierarchical framework laid the foundation for AlexNet (\ncite{krizhevsky_imagenet_2012}), a breakthrough model that set a new standard in visual recognition by using deep architectures to automatically learn complex visual features. AlexNet’s success fueled the development of more advanced models, such as VGGNet (\ncite{simonyan_very_2015}), ResNet (\ncite{he_deep_2016}), and Inception (\ncite{szegedy_going_2015}), whose multi-layered designs mirrored the brain's hierarchical processing stages. 

The brain’s visual system also follows a hierarchical organization, progressing from simple to complex representations. Neurons in early visual areas, such as V1, respond to basic features like edges, while higher regions, such as the inferior temporal cortex, show selectivity for more complex stimuli like faces and objects (\ncite{felleman_distributed_1991}). This quantitative similarity between the hierarchical processing in DNN and the brain’s visual system formed the foundation for the emergence of {\it NeuroAI} \footnote{\textbf{NeuroAI:} An interdisciplinary field combining neuroscience and AI to understand brain and AI functions by integrating neural data with AI representations.}, a field focused on bridging artificial and biological neural networks (\ncite{macpherson_natural_2021}; \ncite{doerig_neuroconnectionist_2023}).

NeuroAI goes beyond mere inspiration. Converging evidence from various approaches demonstrates a more profound similarity between DNN and the brain's visual processing. For example, neural encoding models use DNN-derived features to predict neural responses to visual stimuli, revealing correspondences between specific DNN layers and brain regions (\ncite{yamins_performance-optimized_2014}). Representational similarity analysis (RSA) further supports this alignment by comparing similarity matrices derived from brain activity patterns with those derived from DNN unit activations (\ncite{khaligh-razavi_deep_2014}). These methods have shown that mid-level DNN layers correspond to mid-level visual areas like V4, while higher layers align with higher-level regions, such as the inferior temporal cortex, underscoring DNN's potential as computational proxies for visual processing.

In addition to DNN-based visual recognition models like AlexNet, recent generative AI models, such as GANs (\ncite{goodfellow_generative_2014}) and diffusion models (\ncite{ho_denoising_2020}), have enabled image synthesis that captures intricate details of natural images. These models’ ability to produce photorealistic images makes them promising tools for reconstructing visual experiences from brain activity as well.

Visual recognition and generative models\footnote{\textbf{Generative models:} AI models, including GANs (Generative Adversarial Networks) and diffusion models, that create realistic images or data by transforming latent representations.} form a two-step pipeline that underpins our visual image reconstruction methods, as described in the following section (Figure \ref{fig:image_reconstruction_pipeline}). This pipeline begins with a {\it translator} that maps brain activity to latent representations within a pre-trained DNN-based recognition model. Next, a {\it generator} synthesizes visual images from these predicted (decoded) latent representations using a generative model. This approach enables a seamless transformation from neural signals to high-fidelity visual outputs, effectively bridging the gap between neural and computational representations.

\section{Deep image reconstruction}
The early work by \cite{miyawaki_visual_2008} utilized contrast-based local image bases to decode visual images from brain activity. These bases captured basic visual local contrast features, reflecting the representations at lower layers of DNNs. They were limited in scope, representing only fundamental aspects of visual perception. In contrast, DNN provides a hierarchical structure that parallels the human visual system, capturing different levels of detail, from simple features to complex objects. While intermediate feature representations are often {\it nameless} and {\it faceless}, making them challenging to interpret with simple parametric models or semantics, this alignment allows for a more accurate translation of brain activity into DNN activations, enhancing our understanding of visual processing.

\begin{figure}[htbp]
\centering
\includegraphics[scale=0.62]{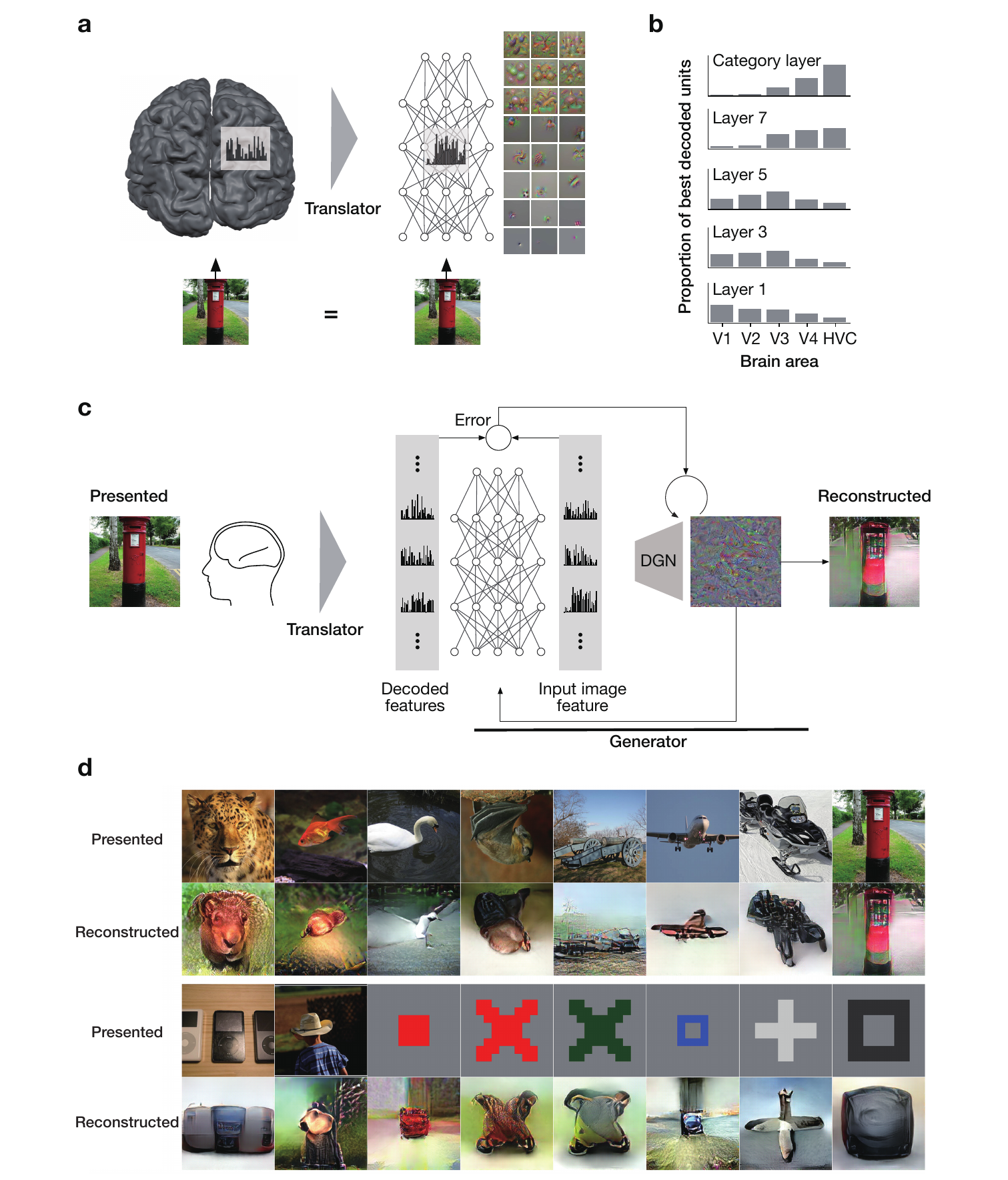}
\caption{Visual image reconstruction using DNN latent representations. (a) Decoding extends from local linear bases to hierarchical DNN features, enabling the translation of hierarchical brain responses into DNN representations while preserving encoded content (\ncite{horikawa_generic_2017}). Brain activity from natural images is used to train DNN feature decoders, pairing neural inputs with DNN unit activations. (b) The alignment between the brain and DNN layers is shown by the distribution of best-decoded units in each DNN layer across visual areas (V1, V2, V3, V4, and higher visual cortex [HVC]). Lower/higher DNN layers align with lower/higher brain regions, respectively. (c) The deep image reconstruction pipeline translates brain activity into DNN features, followed by image generation through image optimization to match DNN feature outputs (Shen et al., 2019b). (d) Reconstructed examples, including both natural and artificial stimuli, demonstrate the model’s generalization to unseen shapes. Adapted from \cite{horikawa_generic_2017}, \cite{shen_deep_2019}, and \cite{nonaka_brain_2021}.}
\label{fig:image_reconstruction_dnn}
\end{figure}

\cite{horikawa_generic_2017} introduced a method that mapped brain activity patterns measured by fMRI onto the latent feature space of a pre-trained DNN (Figure \ref{fig:image_reconstruction_dnn}a). Their model established hierarchical correspondence: Early visual areas corresponded to lower DNN layers representing basic features like orientation, while higher regions aligned with higher DNN layers encoding complex shapes and semantic categories (Figure \ref{fig:image_reconstruction_dnn}b). This work demonstrated functional similarities between DNN and the human visual system from the decoding perspective, emphasizing how DNNs replicate key aspects of biological visual processing. Further research by \cite{horikawa_characterization_2019} showed that the unit-wise correspondence was consistent across individuals, confirming its biological significance. 

Building on these studies, \cite{shen_deep_2019} developed an inverse CNN (iCNN) model that serves as a generator (Figure \ref{fig:image_reconstruction_dnn}c). The iCNN refines image reconstructions by iteratively optimizing an input image to match the brain-decoded latent features, bridging the gap between latent features and final visual outputs. This method effectively improved reconstruction fidelity, generating images that closely resemble the original stimuli observed by subjects (Figure \ref{fig:image_reconstruction_dnn}d). Notably, test images were taken from the categories distinct from the training set, including out-of-domain artificial shapes as in \cite{miyawaki_visual_2008}. Although trained exclusively on natural images, the model demonstrated reasonable accuracy in reconstructing artificial shapes, suggesting that DNN-derived latent features can, albeit imperfectly, represent diverse perceptual experiences.

\begin{figure}[t]
\centering
\includegraphics[scale=0.64]{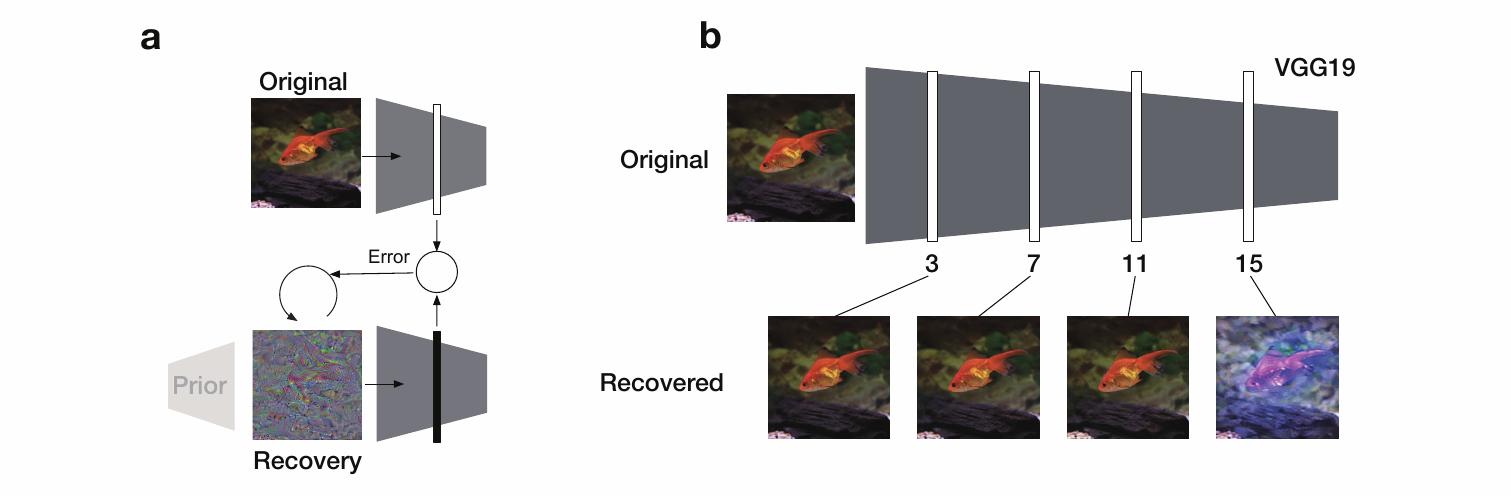}
\caption{Image Recovery from DNN layers. (a) Recovery procedure. DNN features of an image are extracted from a specific layer, followed by an optimization process to generate a new image whose DNN features closely match the extracted features of the original. A weak image prior, such as a deep image prior (\cite{ulyanov_deep_2020}), can refine this process by optimizing both the model parameters and latent features without training on actual images. (b) Examples of recovered images from different layers of the VGG19 model (layers 3, 7, 11, and 15 out of 19) demonstrate that perceptually similar images to the original can be generated even from higher layers. Adapted from \cite{shirakawa_spurious_2024}.}
\label{fig:image_recovery}
\end{figure}

Achieving arbitrary image reconstruction requires compositional latent features that can be effectively mapped to the image space. DNN features often serve as these representations. Contrary to common assumptions, hierarchical processing in DNNs does not discard all pixel-level details. \cite{mahendran_understanding_2015} demonstrated that even mid-level DNN layers preserve sufficient information to reconstruct perceptually similar images. In neuroscience, it has been argued that large receptive fields do not necessarily reduce neural coding capacity if the unit density remains sufficient (\ncite{zhang_neuronal_1999}; \ncite{majima_position_2017}). This suggests that mid- and higher-level DNN layers can support detailed perceptual reconstructions. For instance, by applying a weak image prior based on the deep image prior method (\ncite{ulyanov_deep_2020}), which does not require training on image data, it is possible to recover perceptually similar images even from relatively high DNN layers (Figure \ref{fig:image_recovery}). This demonstrates that DNN-derived latent features can not only represent abstract concepts but also support accurate perceptual reconstructions.

\section{Key components}
Recent developments in visual image reconstruction have expanded technical approaches, paving the way for improvements. This process involves an integrated pipeline designed to decode neural signals into interpretable representations. Despite this progress, challenges such as model misspecification, limited data diversity, output dimension collapse\footnote{\textbf{Output dimension collapse:}  A limitation where model predictions are confined to a subspace, often due to insufficient data or model misspecification.}, and biased evaluation continue to impact model reliability. These issues can lead to {\it spurious reconstructions}, where outputs, though visually convincing, fail to accurately reflect visual perception or generalize beyond the training set. Instead, the outputs may represent mere image retrieval or hallucinations triggered by semantic prompts, and in some cases, visually convincing images may even be generated from random brain data (\ncite{shirakawa_spurious_2024}).

These pipelines can be understood within a {\it translator–generator} framework, although some models integrate these steps end-to-end, mapping brain activity directly to images (e.g., \ncite{shen_end--end_2019}; \ncite{fujiwara_modular_2013}). Nonetheless, the translator–generator model remains valuable for identifying limitations and guiding improvements. Here, this framework will be used to review key components, highlight potential pitfalls, and suggest strategies to enhance model robustness.

\subsection{Latent representation}
Latent representations serve as the critical bridge between neural signals and reconstructed images. However, model misspecification poses a significant challenge, as errors in design can distort the mapping of neural signals to latent features, ultimately leading to inaccurate reconstructions.

A core concept in NeuroAI is the alignment between the hierarchical representations in deep neural networks (DNNs) and human visual processing (\ncite{yamins_using_2016}; \ncite{khaligh-razavi_deep_2014}; \ncite{horikawa_generic_2017}; \ncite{schrimpf_integrative_2020}). While models like AlexNet (\ncite{krizhevsky_imagenet_2012}) and VGG (\ncite{simonyan_very_2015}) have demonstrated similarities to early and higher-level visual processing, this alignment is not consistently observed across all vision models. \cite{nonaka_brain_2021} found that only a limited subset of DNNs—primarily early feedforward models—accurately reflected hierarchical visual representations. This underscores the need for models that more closely mimic human visual processing to enable a more precise translation of neural signals into latent features.

NeuroAI approaches have often relied on neural encoding accuracies of DNN features to evaluate the similarity to brain processing. However, recent critiques suggest that this metric is insufficient, as it does not fully capture the complexity of brain–AI relationships (\ncite{sexton_reassessing_2022}; \ncite{bowers_deep_2022}; \ncite{conwell_what_2023}). To address these shortcomings, researchers are calling for more nuanced evaluations that consider how well models reflect actual neural processes beyond basic encoding accuracies (\ncite{prince_representation_2024}).

Models can leverage either single or multiple DNN layers, customized to suit the perceptual space being reconstructed. Both supervised models, trained for tasks like object recognition (\ncite{simonyan_very_2015}), and unsupervised models, such as autoencoders (\ncite{kingma_auto-encoding_2014}; \ncite{child_very_2021}), can provide meaningful latent representations. Defining the scope of the latent space is critical: while models may focus on specific domains, such as handwritten characters (\ncite{lecun_mnist_2005}; \ncite{schoenmakers_linear_2013}) or faces (\ncite{liu_deep_2015}; \ncite{gucluturk_reconstructing_2017}; \ncite{dado_hyperrealistic_2022}; \ncite{dado_brain2gan_2024}), these limitations must be clearly communicated to avoid overstating the model’s capabilities. Broader claims about reconstructing perception in general require models trained on diverse datasets that encompass a wide range of perceptual experiences.

Integrating multiple modalities, such as visual and semantic features, might enrich latent representations (\ncite{ozcelik_natural_2023}) but must be handled carefully to prevent spurious reconstructions (\ncite{shirakawa_spurious_2024}). Recovery checks are critical to ensure latent features retain the necessary perceptual details. 

\subsection{Translator}
The translator component is responsible for mapping brain signals to latent features. Historically, linear models have been preferred for this task due to their simplicity and robustness, especially when dealing with limited training samples (\ncite{kamitani_decoding_2005}; \ncite{yamashita_sparse_2008}; \ncite{miyawaki_visual_2008}). These models help prevent overfitting, reducing the risk of falsely attributing neural encoding to spurious model features (\ncite{kamitani_decoding_2012}).

However, linear models often face output dimension collapse, where shared input variables cause dependencies among output features, limiting the diversity of outputs (\ncite{shirakawa_spurious_2024}). This constraint hinders compositional predictions, which are essential for accurately reconstructing complex visual scenes. Furthermore, a lack of dataset diversity can exacerbate this collapse, reducing the model's ability to generalize across different types of stimuli.

To address these limitations, models designed to encourage input variable sparsity have been developed (\ncite{yamashita_sparse_2008}; \ncite{fujiwara_modular_2013}). When combined with appropriate regularization, these models ensure that each feature is predicted more independently, maintaining diversity in latent spaces. Input feature selection techniques have also been explored to create more modular decoders (\ncite{shen_deep_2019}). By focusing on relevant neural features, these modular decoders enhance generalizability and accommodate a broader range of latent representations, which is crucial for reconstructing complex and novel visual stimuli.

\subsection{Generator}
The generator is responsible for transforming latent features into visually interpretable images. Various methods have been developed for this purpose, each varying in complexity and output fidelity. The early pixel optimization approach (iCNN; \ncite{shen_deep_2019}) iteratively refined pixel values to match decoded latent features. This process was improved by a deep generative network (DGN) based on GAN, which is trained to generate natural images and act as an image prior. This approach enabled optimization in a lower-dimensional input space, enhancing both efficiency and output quality.

Recent methods have shifted toward direct generators that bypass iterative optimization altogether. Models like GANs and diffusion models can be used to generate outputs directly from latent features (\ncite{cheng_reconstructing_2023}; \ncite{dado_brain2gan_2024}; \ncite{ozcelik_natural_2023}). While GANs produce visually coherent images, diffusion models offer even greater flexibility, generating diverse outputs from the same neural data (\ncite{cheng_reconstructing_2023}). This flexibility, however, introduces potential concerns around cherry-picking, where researchers may selectively present only the most visually appealing results, potentially compromising scientific integrity and reproducibility. To ensure rigor, it is essential to apply transparent selection criteria and avoid conflating a realistic appearance with accurate perceptual content (\ncite{shirakawa_spurious_2024}).

\subsection{Dataset}
Dataset diversity is critical for developing robust and generalizable models of visual image reconstruction. A lack of diversity in training datasets can lead to model bias, limiting the ability to generalize to unseen stimuli. This deficiency often results from output dimension collapse, where the latent features predicted from brain activity become restricted to a narrow subspace, thereby reducing the fidelity of reconstructed percepts (\ncite{shirakawa_spurious_2024}).

Standardized datasets, such as the Natural Scene Dataset (NSD; \ncite{allen_massive_2022}), have provided valuable benchmarks for model evaluation. However, the limited scope of such datasets can introduce biases, restricting model generalizability and influencing reconstruction outcomes. Even when test images are not directly included in the training, foundational models like CLIP (\ncite{radford_learning_2021}) may have been trained on similar data, potentially skewing results. This overlap raises concerns about latent representations inadvertently reflecting test stimuli, which can compromise the validity of the reconstructions.

Researchers should exercise caution when combining pre-trained models with publicly available datasets. Careful selection and validation are essential to prevent biases and to ensure that models genuinely reflect the neural encoding of diverse perceptual experiences. A lack of dataset diversity increases the risk of overfitting to specific categories, limiting the model’s ability to accurately reconstruct a broad range of visual stimuli.

\subsection{Evaluation}
Proper evaluation is crucial for ensuring the reliability of visual image reconstruction models. Key practices include avoiding cherry-picking and ensuring generalizability beyond the training images (\ncite{shirakawa_spurious_2024}). Effective evaluation requires true zero-shot prediction, where models generate outputs for previously unseen stimuli. The problem of using the same or similar images in both training and testing is related to but distinct from double-dipping, which involves reusing the same data including both input and output variables, particularly in feature selection (\ncite{kriegeskorte_circular_2009}).To achieve genuine generalization, it is essential to confirm that test stimuli were not included in any phase of model training.

A widely used metric in model evaluation is pairwise identification accuracy, which assesses how well-predicted features or images match the true features or images in a pairwise comparison within a test set (\ncite{shen_end--end_2019}; \ncite{shen_deep_2019}; \ncite{ozcelik_natural_2023}; \ncite{scotti_reconstructing_2023}). However, this metric has limitations: it can yield high scores even when models capture only broad categories rather than finer perceptual details (\ncite{shirakawa_spurious_2024}). While quantitative metrics offer objective assessments, they cannot fully capture the perceptual accuracy of reconstructions. Therefore, qualitative evaluations must confirm that reconstructions genuinely reflect the original perceptual experiences, not just category-level similarities.

\subsection{Inter-Individual Analysis}
Inter-individual analysis in visual image reconstruction presents significant challenges due to the variability in neural responses across individuals. Traditionally, reconstruction models rely on training and testing data from the same individuals (\ncite{miyawaki_visual_2008}; \ncite{shen_deep_2019}), which limits generalizability and restricts the model's applicability to broader populations. This within-subject approach results from the fine-grained, individualized nature of neural coding, making cross-subject reconstructions particularly difficult.

Recent developments have sought to overcome these challenges through techniques for functional alignment, which aligns neural responses across individuals (\ncite{haxby_hyperalignment_2020}). Several techniques have shown promise in achieving cross-subject reconstruction (\ncite{yamada_inter-subject_2015}; \ncite{ho_inter-individual_2023}; \ncite{wang_inter-individual_2024}). These methods enhance the flexibility of inter-individual models, expanding the potential applications of brain decoding across diverse populations, with implications for clinical applications and brain-machine interfaces.

\section{Image reconstruction as psychological measurement}
Psychological measurement involves systematically quantifying mental processes by assigning measurable variables to them. Traditionally, this has been achieved through psychophysical methods that rely on simple behavioral responses, such as reaction times or subjective ratings, to infer perceptual experiences (\ncite{kingdom_psychophysics_2016}). Neuroimaging approaches, on the other hand, have focused on identifying {\it biomarkers} of psychological processes, typically capturing coarse representations of mental states. In contrast, visual image reconstruction could offer a novel method by transforming internal mental states into measurable outputs composed of pixel values. 

Fechner’s {\it inner psychophysics} concept is a foundational framework for understanding how neural activity corresponds to subjective experiences (\ncite{fechner_elements_1966}; \ncite{ritchie_neural_2016}; \ncite{cheng_deciphering_2024}). In contrast to the conventional psychophysics ({\it outer psychophysics}), which links physical stimuli to sensory responses, inner psychophysics examines the neural correlates of mental states. Fechner envisioned this as a theory of the precise mapping between mental quantities and brain processes. Although limited by the tools of his time, Fechner’s vision has been partially realized through modern brain decoding, which aims to establish a direct correspondence between neural signals and subjective percepts (\ncite{ritchie_neural_2016}; \ncite{kamitani_decoding_2005}; \ncite{naselaris_voxel-wise_2015}). Image reconstruction, as a form of inner psychophysics, thus represents a crucial step toward translating neural activity into concrete representations of subjective mental content.

\begin{figure}[htbp]
\centering
\includegraphics[scale=0.6]{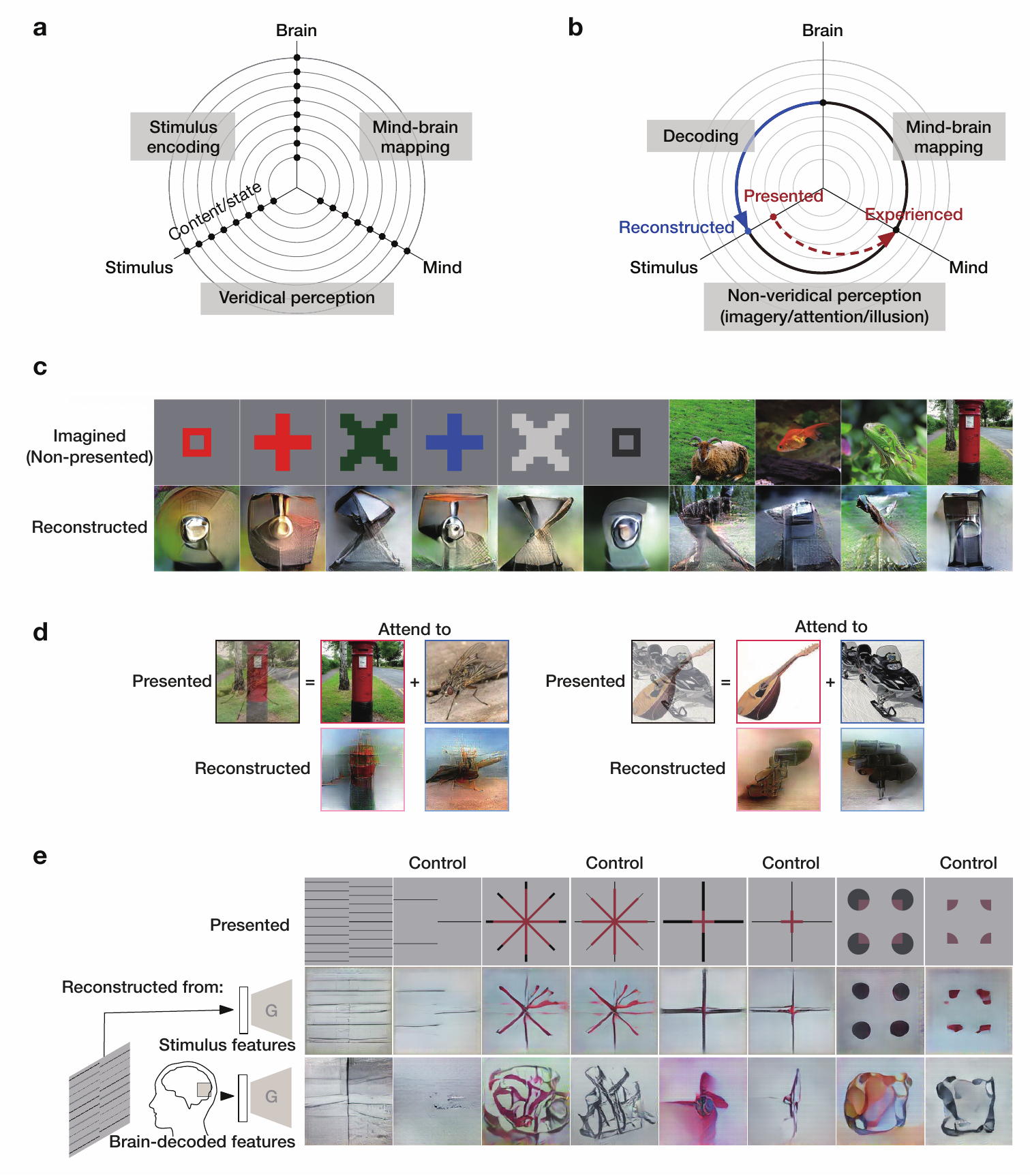}
\caption{Psychological measurement of subjective visual experiences through image reconstruction. (a) Mapping of brain, stimulus, and mind. Dots represent instances of visual experience (e.g., an image, perception, and corresponding brain activity). Veridical perception assumes that the mind accurately represents stimuli. The brain–mind mapping is considered fixed, while the brain–stimulus relationship is empirically identified. (b) Non-veridical perception (e.g., mental imagery, attentional modulation, and illusions) occurs when perceived content diverges from physical properties. The fixed brain–mind mapping and decoders trained on brain activity under veridical conditions allow the reconstruction of mental content as an image. (c) Reconstruction of mental imagery is achieved using models trained on brain activity from natural images (\ncite{shen_deep_2019}). (d) Reconstruction under selective attention highlights the attended image among overlapping stimuli (\ncite{horikawa_attention_2022}). (e) Reconstructions with illusory stimuli and the control conditions (illusory line and color spreading), along with reconstructions from the stimulus features (\ncite{cheng_reconstructing_2023}) reveal that illusory effects emerge only when using brain-decoded features, ensuring reconstructions reflect brain activity specific to illusory experiences rather than model artifacts. Adapted from \cite{shen_deep_2019}, \cite{horikawa_attention_2022}, and \cite{cheng_reconstructing_2023}.}
\label{fig:psychological_measurement}
\end{figure}

Beyond reconstructing perceptions, image reconstruction techniques have extended to representing subjective experiences, such as mental imagery, selective attention, and perceptual illusions. The central assumption is that neural representations are at least partly shared between externally driven perception and internally generated mental states when the content aligns (\ncite{kamitani_decoding_2005}). This shared representational framework allows models trained on stimulus-driven neural activity to decode subjective experiences (Figure \ref{fig:psychological_measurement}a,b). 

In veridical perception\footnote{\textbf{Veridical perception:} The accurate perception of stimuli as they exist in the external world, without distortion from subjective interpretation.}, where the mind is assumed to closely represent external stimuli, the instances of stimulus and mental content generally align, and both can be linked to the same brain state (Figure \ref{fig:psychological_measurement}a). However, it is important to recognize that perfect veridical perception is an idealization, achievable only approximately. Leveraging brain activity patterns under veridical conditions, a decoder can connect stimulus content, mental content, and brain state. In non-veridical perception, such as mental imagery, selective attention, and perceptual illusions, the presented stimulus does not align with the experienced content (or, as with imagery, there can be no external stimulus at all). Nevertheless, a decoder trained on veridical perception can map the brain state to a stimulus that corresponds with the experience. This is possible because the mapping between the brain and the mind is assumed to be shared between stimulus-induced perception and internally generated subjective experiences. Through this process, subjective mental content could be externalized from brain activity into an image that mirrors the internal experience. 

\cite{shen_deep_2019} demonstrated this approach in a study on mental imagery (\ncite{kosslyn_neural_2001}). They used the same decoder trained for reconstructing perception to examine mental imagery. Before fMRI scans, subjects memorized test images; later, during scans, they were asked to imagine these images. The perception-trained reconstruction model showed that imagery-related neural activity could be transformed into visual outputs that roughly matched the original stimuli, especially for simple shapes (Figure \ref{fig:psychological_measurement}c). More complex images were less accurately reconstructed, indicating a possible discrepancy between neural representations of perception and imagery.

\cite{horikawa_attention_2022} similarly investigated how attentional modulation influences visual perception. By training a model on stimulus-driven brain activity, they applied it to neural data collected during selective attention tasks, where subjects focused on one of two overlapping images. The resulting reconstructions emphasized features of the attended image, demonstrating that attention can shape the neural representation of the same stimulus (Figure \ref{fig:psychological_measurement}d). This finding illustrates that a single visual stimulus can produce different reconstructions depending on the observer’s attentional state, thus externalizing subjective focus.

Illusory perceptions offer a valuable demonstration of how this reconstruction framework operates. \cite{cheng_reconstructing_2023} applied a model trained on veridical perception to decode neural responses to visual illusions, including illusory lines and neon color spreading. Despite the absence of these features in the true stimulus, the reconstructions captured the illusory content as perceived by subjects (Figure \ref{fig:psychological_measurement}e). This finding indicates that the brain’s internal interpretation of visual inputs during illusions can be externalized, blending both true stimulus features and subjective experiences into a unified visual output. Importantly, their approach verified that the decoder did not inherently produce illusory elements, as reconstructions based solely on stimulus features—without brain input—lacked illusory content. This distinction underscores that illusory features emerge specifically through brain-decoded representations, highlighting the decoder’s ability to map subjective experience.

Additionally, these studies compared reconstruction accuracy across different visual areas, revealing variable success rates in capturing subjective experience versus purely stimulus-driven perception. Such findings indicate which brain areas contribute to subjective experience, stimulus-induced perception, or both, providing insights into how mental representations are generated in the brain. For example, early visual areas may excel in decoding direct stimulus features, while higher areas often more accurately capture attention-modulated or illusory content. This comparative analysis offers valuable information on the dynamic processes underlying mental representation in the brain.

While this approach assumes a shared representational space between perception and subjective experience, this assumption may not always hold (\ncite{breedlove_generative_2020}; \ncite{cabbai_sensory_2024}). Differences in top-down versus bottom-up processing, for example, may influence how the contents are represented in the brain. Additionally, neural variability introduced by individual history or cognitive states may impact reconstruction accuracy. To address these challenges, models could be adapted to include brain activity data from subjective tasks, such as voluntary imagery, to enhance internal visual generation. Furthermore, integrating multimodal data that captures contextual and emotional factors could improve decoding accuracy for subjective experiences.

\section{Ethical considerations}
The rapid advancement of neurotechnology has spurred considerable ethical debate, particularly concerning privacy and the protection of cognitive liberty (\ncite{yuste_four_2017}; \ncite{greely_neuroethics_2019}; \ncite{unesco_unveiling_2023}). Brain decoding, often framed in popular media as {\it mind-reading}, has captured public attention due to its potential applications in fields such as medicine and brain-machine/computer interfaces (BMIs). However, this heightened interest also raises critical ethical questions about how brain data should be used, accessed, and stored. As image reconstruction and brain decoding techniques advance, they challenge our concepts of mental privacy and highlight the need for balanced, informed regulation (\ncite{ienca_towards_2017}).

Privacy is a central concern in the ethics of brain decoding technology, given the potential to access sensitive personal information without the subject’s full awareness. For instance, \cite{horikawa_neural_2013} demonstrated that dream content could be semantically decoded from brain activity, potentially revealing mental experiences that subjects may not consciously remember. While current technology cannot reconstruct detailed visual imagery from dreams, studies like this underscore neurotechnology’s potential to access deeply personal information. Another area of concern is brain-computer interfaces (BCIs), which use neural data from implanted electrodes to interpret or communicate visual and semantic information (e.g., \ncite{fukuma_voluntary_2022}). This technology line may carry ethical risks associated with inadvertently accessing private thoughts through neural data.

However, it is essential to frame these concerns within the realistic limitations of current technology. Presently, visual reconstruction methods depend heavily on subject cooperation, such as remaining still in the scanner or interface device, and many decoding models require extensive training on each subject to achieve reliable results (\ncite{tang_semantic_2023}). Additionally, test data are often produced by averaging multiple trials to enhance signal clarity, further limiting real-time or single-trial decoding accuracy. These limitations make unauthorized or involuntary decoding implausible at present. Yet, the public often perceives these technologies as more advanced than they are due to exaggerations in the media (\ncite{farah_neuroethics_2012}). Misinformation about {\it mind-reading capabilities} can lead to premature regulation by policymakers and legal experts who may lack a comprehensive understanding of the technology’s limitations, potentially stifling valuable research and development.

\section{Conclusions}
Brain decoding has advanced rapidly, evolving from basic classification tasks to complex reconstructions that reflect subjective experience. This progress underscores the importance of decoding, especially when combined with DNN. While these models produce increasingly realistic reconstructions, challenges remain, emphasizing the need for a balanced approach that incorporates broader neuroscientific methods.

In neuroscience, encoding models have traditionally dominated for their structured approach to linking stimuli with neural responses, as seen in recent NeuroAI studies where DNN features are used as input to predict brain activity (\ncite{yamins_performance-optimized_2014}). However, dismissing decoding as a secondary {\it parlor trick} overlooks its value (\ncite{vigotsky_mental_2024}). As early neural coding studies demonstrated, encoding and decoding are fundamentally interconnected (\ncite{bialek_reading_1991}; \ncite{rieke_spikes:_1997}). Decoding produces tangible outputs that allow researchers to define meaningful effects, which are often difficult to interpret in predicted neural signals alone.

From a psychological perspective, brain decoding supports Fechner’s concept of inner psychophysics, linking neural data with subjective experience (\ncite{fechner_elements_1966}; \ncite{ritchie_neural_2016}). This provides new avenues in psychological measurement, expanding on classical psychophysical approaches by incorporating neural data into perceptual models.

Although DNN-based decoding relies on {\it nameless} and {\it faceless} features rather than interpretable units, this does not diminish its value, as the brain itself may not be inherently interpretable (\ncite{nastase_keep_2020}; \ncite{hasson_direct_2020}). It remains essential to examine DNN-brain alignment on a broader scale without presuming interpretability at the level of individual units. In our approach, DNNs serve as {\it feature generators} rather than direct neural analogs. While mechanistic insights through DNNs may be promising, it is crucial to remember that the developmental and operational dynamics of the visual cortex diverge significantly from current DNN training processes, reinforcing DNNs’ role as complementary tools rather than comprehensive brain models (\ncite{murakami_modular_2022}).

In conclusion, advancing visual image reconstruction demands rigorous, ethically grounded research. A balanced integration of AI models with neuroscience and psychology will be essential for meaningful progress in this evolving field.

\section{Summary Points}
\begin{enumerate}
    \setlength{\itemsep}{0.5em}
    \item Brain decoding of visual perception has evolved from basic classification tasks to sophisticated techniques capable of generating complex images that approximate subjective visual experiences from brain activity.
    \item Deep neural networks (DNNs) and generative models have been instrumental in enhancing the fidelity of reconstructions by leveraging hierarchical latent representations of visual information.
    \item Achieving true zero-shot prediction remains a critical challenge, as current models often struggle to generalize beyond their training data, limiting their applicability to novel visual inputs.
    \item Compositional representations, which treat complex images as combinations of simpler elements, show promise for improving model flexibility and generalization.
    \item Datasets with diverse contents are essential for training and testing models for robustness and generalizability.
    \item Spurious reconstructions can occur when models generate realistic but inaccurate outputs, emphasizing the need for rigorous evaluation to ensure reconstruction authenticity.
    \item Generalizing decoders from stimulus-induced to subjective conditions enables the reconstruction of subjective contents, serving as a novel approach to psychophysical measurement.
    \item Ethical considerations, including privacy, consent, and potential misuse, are central to the responsible development of brain decoding technologies, while researchers should avoid presenting overly optimistic claims based on spurious reconstructions.
\end{enumerate}

\section{Future Issues}
\begin{enumerate}
    \setlength{\itemsep}{0.5em}
    \item Developing DNNs with brain-like representations, along with metrics to evaluate their {\it brain-likeness}, will enhance the relevance and interpretability of these models in neuroscience applications.
    \item Incorporating top-down influences, such as attention, prior knowledge, and context, will be crucial for accurately capturing subjective visual experiences in reconstructions.
    \item Expanding datasets to include a broader range of visual elements is necessary for training reconstruction models and assessing their generalization performance across diverse visual experiences.
    \item Developing advanced evaluation metrics that align with human perceptual judgments will help quantify reconstruction quality and complement subjective ratings.
    \item Research should explore clinical applications of reconstruction technologies, including personalized diagnostics and therapeutic interventions for visual or cognitive impairments.
    \item Creating real-time reconstruction systems will enable dynamic visualization of brain activity, facilitating interactive brain-computer interfaces and new experimental paradigms.
    \item Pooling data across individuals through functional alignment is essential for scalable analysis and for training models that generalize effectively across diverse individuals.
\end{enumerate}

\noindent
\setlength{\fboxsep}{10pt}
\fbox{%
  \parbox{\dimexpr\linewidth-3\fboxsep-3\fboxrule}{%
    \textbf{Neuroaesthetics and brain-generated art} \vspace{0.5em} \\
    The emerging field of neuroaesthetics examines how the brain processes aesthetic experiences, such as those evoked by art, music, and beauty. As visual image reconstruction techniques advance, they increasingly intersect with neuroaesthetics, enabling the visualization of subjective responses to art directly from brain activity. This progress not only sheds light on individual aesthetic preferences but also facilitates the creation of brain-generated art. Artists and neuroscientists are now collaborating to transform neural reconstructions into creative works, demonstrating neurotechnology’s potential to bridge science and art. Projects like UUmwelt by Pierre Huyghe at the Serpentine Galleries and Liminal at Punta della Dogana use reconstructions from brain data to explore themes of externalizing and sharing consciousness, offering a novel medium for artistic expression. This intersection underscores how brain decoding can illuminate perceptual and cognitive processes while contributing to a deeper understanding of the neural basis of creativity and artistic experience.
  }
}

\section{DISCLOSURE STATEMENT}
YK is a part-time researcher at Advanced Telecommunications Research International (ATR), which holds patents related to some of the work discussed in this review.

\section{ACKNOWLEDGMENTS}
We are grateful to the current and former members of the Kamitani Lab for their valuable discussions and insights, which have contributed to the development of this review. This work was partially supported by JSPS KAKENHI (20H05705 and 20H05954), NEDO (JPNP20006), JST CREST (JPMJCR22P3), and AMED (JP24wm0625409). The funders had no role in the study design, data collection, analysis, decision to publish, or preparation of the manuscript.

\vspace{1em}

\noindent\textbf{Related Resources}

\hangindent=1em
\hangafter=1
\noindent
Kamitani Lab GitHub: A repository offering code, data, and tools from Kamitani Lab that supports research in brain decoding and visual image reconstruction. \\
\url{https://github.com/KamitaniLab} 

\hangindent=1em
\hangafter=1
Kamitani Lab YouTube Channel: A collection of videos showcasing the lab’s brain-generated contents using brain decoding and visual image reconstruction techniques.  \\
\url{https://youtube.com/@atrdni?si=lOu35br6HZQYseOW}

\hangindent=1em
\hangafter=1
UUmwelt at Serpentine Galleries: An art installation featuring visual image reconstructions from Kamitani Lab, which explores themes at the intersection of art, neuroscience, and AI.  \\\url{https://www.serpentinegalleries.org/whats-on/pierre-huyghe-uumwelt/}

\hangindent=1em
\hangafter=1
Liminal at Punta della Dogana: This exhibition uses brain-generated imagery to investigate the boundaries of perception and reality through visual reconstructions.  \\\url{https://www.pinaultcollection.com/palazzograssi/en/pierre-huyghe-liminal}

\hangindent=1em
\hangafter=1
UNESCO’s Neurotechnology Ethics: A resource addressing ethical considerations in neurotechnology, including privacy, human rights, and the responsible use of brain data.  \\
\url{https://www.unesco.org/en/ethics-neurotech}

\vspace{0.5em}

\end{document}